\documentclass[runningheads]{llncs}
\usepackage{graphicx}
\usepackage{booktabs}
\usepackage{tabularx}
\usepackage{hyperref}
\usepackage{paralist}
\usepackage{tabulary}
\usepackage{ragged2e}
\usepackage{wrapfig}
\usepackage{cite}
\usepackage{comment}
\usepackage{caption}

\newcolumntype{Y}{>{\centering\arraybackslash}X}

\begin{document}
\title{A Reference Data Model for Process-Related User Interaction Logs}
%
%
\author{Luka Abb\orcidID{0000-0002-4263-8438} \and
Jana-Rebecca Rehse\orcidID{0000-0001-5707-6944}}
\authorrunning{Abb and Rehse}
%
\institute{University of Mannheim \\
\email{\{luka.abb,rehse\}@uni-mannheim.de}}
\maketitle              
\begin{abstract}
User interaction (UI) logs are high-resolution event logs that record low-level activities performed by a user during the execution of a task in an information system. Each event in a UI log corresponds to a single interaction between the user and the interface, such as clicking a button or entering a string into a text field. UI logs are used for purposes like task mining or robotic process automation (RPA), but each study and tool relies on a different conceptualization and implementation of the elements and attributes that constitute user interactions. This lack of standardization makes it difficult to integrate UI logs from different sources and to combine tools for UI data collection with downstream analytics or automation solutions. To address this, we propose a universally applicable reference data model for process-related UI logs. Based on a review of scientific literature and industry solutions, this model includes the core attributes of UI logs, but remains flexible with regard to the scope, level of abstraction, and case notion. We provide an implementation of the model as an extension to the XES interchange standard for event logs and demonstrate its practical applicability in a real-life RPA scenario.

\keywords{User Behavior Mining \and UI Log \and Data Model \and Robotic Process Automation \and Task Mining} 
\end{abstract}
\section{Introduction}

\enlargethispage{\baselineskip}

\emph{User interaction (UI) logs} are high-resolution event logs that record low-level, manual activities performed by a user during the execution of a task in an information system (IS) \cite{abb_2022_ubm}. Each event in a UI log corresponds to a single interaction between the user and the graphical user interface (GUI) of a software application. Examples include clicking a button, entering a string into a text field, ticking a checkbox, or selecting an item from a dropdown \cite{LenoD1}.
Multiple recent research streams use this type of data, 
for example to analyze usage patterns in software applications \cite{Dev2017,Linares2015,Ardimento2019}, to identify candidate routines for robotic process automation (RPA) \cite{LenoD1,LenoD3,Urabe2021}, or to derive RPA automation and test scripts \cite{Agostinelli2020,JimenezD1}. In addition, companies like Celonis and UiPath offer tools that record and process UI data for inspecting and automating task executions \cite{AgostinelliD1}. 

The UI logs currently used in research differ substantially. The data collected in a specific research context is usually limited in scope and tailored to the proposed analysis technique or automation approach. This results in considerable variation regarding the number, type, and granularity of recorded events and corresponding attributes. Even when researchers record the same attributes at a similar level of detail, there is no common definition of UI log attributes to which they can adhere. Instead, they often rely on ad-hoc conceptualizations of elementary notions like activities and UI components. The situation is similar in industry, where each vendor has developed their own UI log format tailored to the capabilities of their recording software \cite{Leno2019}.

This lack of standardization makes it difficult to integrate UI logs from different sources \cite{LenoD1,Lopez2020}. It also poses a challenge for the interoperability of data collection and downstream processing tools: logs recorded by one tool are usually only compatible with the associated analytics or automation approach. Combining data collection and processing tools requires considerable preprocessing effort or is entirely infeasible if the necessary attributes cannot be recorded \cite{Leno2019}.

In this paper, we address these challenges by proposing a reference data model for process-related UI logs.
This model provides a data structure and an accompanying interchange format that others can reuse to conceptualize and capture UI logs in a process context.
To ensure widespread applicability, the model is designed such that it subsumes and integrates the commonalities of existing process-related UI logs, but remains flexible with regard to the their differences. 
To identify those commonalities and differences, we conduct a literature review in \autoref{sec:literaturereview} and a review of industry solutions in \autoref{sec:industryreview}. 
The reference data model, along with its underlying design principles and an accompanying interchange format, is presented in \autoref{sec:data_model}. 
In \autoref{sec:evaluation}, we demonstrate how the data model can be instantiated in practice by applying it in a real-life RPA scenario. 
Finally, we conclude the paper with a discussion in \autoref{sec:discussion}.

\section{Background and Related Work}

\enlargethispage{\baselineskip}

\textbf{Event Logs.} Process mining extracts information from \emph{event logs}, i.e., collections of \emph{events} recorded in an IS \cite{TaskMining}. An event log consists of \emph{cases} that each correspond to one process instance. Each case contains a trace of events that occurred during the execution of the process instance and can have additional attributes, for example, the size of an order in an order-to-cash process. Events are related to a particular step in a process with an \emph{activity} label (e.g., create invoice) and can also have additional attributes. 

\noindent
\textbf{Data Formats.} To enable the exchange of event data between different ISs, the business process management (BPM) community has developed interchange formats that define the structure and general contents of event logs. The current main format is XES (eXtensible Event Stream), which was introduced in 2010 to replace the older MXML format and was accepted as the official IEEE standard for event data in 2016 \cite{XES-standard-IEEE}. In XES, an event log consists of a three-level hierarchy of log, trace, and event objects. The format is designed to be highly generic, with a minimal set of explicitly defined attributes on each of the three levels. Additional attributes, with a commonly understood semantic meaning, can be introduced by XES extensions. For example, the concept extension introduces the ``name'' attribute, which stores names for event logs, traces, and events.
Although researchers have recently pointed out shortcomings of XES 
and proposed more flexible, object-centric alternatives such as OCEL \cite{OCEL}, XES remains the most common event log format and is supported by many process mining tools. 

\noindent
\textbf{UI Logs.} UI logs are a particular type of event log in which events correspond to low-level interactions of a user with a GUI. They can be recorded either internally by adding logging capabilities to an application, or externally by dedicated logging tools. These tools record screen coordinates for each action and map them to parts of the GUI using optical character recognition technology.

\noindent
\textbf{User Behavior Mining.} UI logs essentially record how users behave while they are engaged with an application. They can be analyzed by means of data or process mining techniques to gain data-driven insights into user behavior. We refer to this analysis of UI logs as \emph{user behavior mining (UBM)} \cite{abb_2022_ubm}. UBM can serve different purposes, including the analysis of software usage patterns, the design of new user assistance components, or the automation of tasks. 

\noindent
\textbf{Task Mining.} One application of UBM is to enhance traditional process mining by providing a more detailed view of execution steps. Event logs gathered from ERP systems like SAP or Oracle capture the main tasks in a process, like creating an order, but they do not provide insights into how employees actually perform these tasks. Recording and analyzing detailed task executions is referred to as \emph{task mining} \cite{TaskMining} or \emph{desktop activity mining} \cite{Linn2018}. These techniques can give companies deeper insights into their processes than traditional process mining alone, and they can also help software vendors to optimize their products, for example, by identifying common usability issues.

\noindent
\textbf{Robotic Process Automation.} UI logs can be used to automate tasks and entire processes by having bots emulate the recorded user interactions. This approach to automation is called \emph{robotic process automation (RPA)} \cite{Jimenez2019} and has lately received considerable attention in research and practice. Within RPA, UBM techniques can be used to derive automation scripts, but also for \emph{robotic process mining} \cite{LenoD1}, which for example encompasses the identification of suitable tasks for automation from UI logs \cite{LenoD2}.

\noindent
\textbf{Web Usage Mining.} Another field that is concerned with the analysis of user behavior is \emph{web usage mining} \cite{Srivastava2000}, i.e., the analysis of clickstream user data recorded during interactions with websites. Web usage mining is often process-agnostic; its main purpose is to optimize websites, for example, by adapting their content and structure to users' browsing behavior \cite{Ho2010,Ding2015}. The primary data source for web usage mining are server UI logs that are generated in a standardized logging format like the Extended Log Format \cite{extended_log_format} and have a fixed set of attributes.
These include the URL of the current and previous page request, the resource accessed, timestamps, identifying data like the user's IP address, and technical data about the user's web browser and operating system.

\enlargethispage{\baselineskip}

\noindent
\textbf{UBM in other domains.} In addition to the research areas mentioned above, interaction logs have been used as a source of data-driven insights into user behavior in several other domains, such as human-computer-interaction \cite{HCI, HCI2}, information retrieval \cite{Searchlogs}, and visualization \cite{Vislog}. The logs in these domains can take various forms, but they generally record user interactions at a much lower level of detail than the process-related UI logs that we focus on in this paper.

\section{Literature Review} 
\label{sec:literaturereview}

This paper's goal is to develop a reference data model that subsumes and integrates the commonalities of current approaches for capturing process-related UI logs, but stays flexible with regard to their differences. In this section, we review UI logs from scientific literature to identify those commonalities and differences.


\subsection{Research Method}

We conducted a structured literature review \cite{kitchenham2004procedures} in SpringerLink, IEEE Xplore, and ACM Digital Library. As search terms, we used ``log'' combined with \begin{inparaenum}[(1)]
\item ``user interact*'' and ``user interface''
\item ``task mining'' and ``desktop activity mining'' as common terms for high-resolution process mining, and
\item ``robotic process automation'' and ``robotic process mining'' as important applications of UI logs.
\end{inparaenum}
We limited our search to papers written in English and published after 2015 because we focus on the current state of the art.
The relevance of the initial search results was assessed based on their title and abstract. 
This yielded a set of potentially relevant papers, on which we performed a forward-backward-search to also cover papers that our search terms might have missed.

To ascertain the relevance of the identified papers, we scanned their full text for passages on UI logs or recording approaches for them.
Papers were considered as relevant if (1) they contained a concrete UI log or (2) they described the UI log collection process in enough detail to infer the captured attributes. 

\begin{table}[htb]
\vspace{-2em}
\caption{The papers found in the database search}
\begin{tabularx}{\textwidth}{p{.45\textwidth}XXX}
\toprule
\textbf{Search Term} & \textbf{IEEE Xplore} & \textbf{ACM DL} & \textbf{SpringerLink} \\
\midrule
log AND ``user interact*'' & \cite{LenoD2} & \cite{Dev2017} & \cite{AgostinelliD3, Urabe2021, LenoD3, Agostinelli2020} \\
log AND ``user interface'' & & & \\
log AND ``task mining'' & & & \cite{LenoD1} \\
log AND ``desktop activity mining'' & & & \\
log AND ``robotic process automation'' & & & \cite{AgostinelliD1, Jimenez2019} \\
log AND ``robotic process mining'' & & & \\
\bottomrule
\end{tabularx}
\label{tab:reviewfinds}
\end{table}

As listed in \autoref{tab:reviewfinds}, we found 9 relevant publications in the initial search. Several papers appeared in more than one query, but are only listed under the search term that we first found them with.
The forward-backward search returned another 10 publications. 
Although we did not explicitly search for web usage mining logs, our search returned papers about server-side and also client-side web usage mining (recording web activities by adding tracking software to a browser), but none of these met the above-listed criteria. In our review, we therefore only included one exemplary clickstream log from a process mining context: the BPI Challenge 2016 \cite{BPIC2016}, in which the Dutch Employee Insurance Agency recorded eight months of user activities on their website.
Our final result was hence a set of 20 relevant publications.


\subsection{Results}

Some of the 20 relevant publications covered the same use case and data collection approach and were therefore treated as duplicates, resulting in 12 unique approaches. The majority of papers cover RPA \cite{AgostinelliD3,Agostinelli2020,AgostinelliD2,AgostinelliD1,Choi2021,Hofmann2021,JimenezD1,Jimenez2019,LenoD3,LenoD2,LenoD1,Leno2019,LenoD4,Urabe2021}. Four publications \cite{Ardimento2019,Damevski2017,Dev2017, Linares2015} focus on software process mining \cite{Rubin2014}. The remaining two are general approaches to analyzing low-level user interactions with broader applications \cite{BPIC2016,Linn2018}.

\noindent
\textbf{Commonalities.}
Although the reviewed UI logs were fairly heterogeneous, we found a set of six core attributes that are recorded in more than half of them. 
\autoref{tab:litreview} indicates which of the 12 approaches include which attributes ($\bullet$). 

\begin{table}[htb]
\vspace{-2em}
\caption{UI log attributes as found in the literature review}
\begin{tabularx}{\textwidth}{cYYYYYY}
\toprule
\textbf{Source} & \textbf{Action type} & \textbf{Target element} & \textbf{UI Hierarchy} & \textbf{Appli-cation} & \textbf{Input value} & \textbf{Time\-stamp} \\
\midrule
\cite{AgostinelliD3,Agostinelli2020,AgostinelliD2,AgostinelliD1} & $\bullet$ & $\bullet$ & $ \bullet$ & $\bullet$ & $\bullet$ & $\bullet$ \\
\cite{Ardimento2019} & $\bullet$ & $\bullet$ & & single & $\bullet$ & $\bullet$ \\
\cite{Choi2021} & $\bullet$ & $\bullet$ & $ \bullet$ & $\bullet$ & $\bullet$ & $\bullet$ \\
\cite{Damevski2017} & $\bullet$ & $\bullet$ & & single & & $\bullet$ \\
\cite{BPIC2016} &  & $\bullet$ & $\bullet$ & single & $\bullet$ & $\bullet$ \\
\cite{Dev2017} & $\bullet$ & & & single & & $\bullet$ \\
\cite{Hofmann2021} & $\bullet$ & $\bullet$ & $\bullet$ & $\bullet$ & & $\bullet$ \\
\cite{JimenezD1,Jimenez2019} & $\bullet$ & & & $\bullet$ & $\bullet$ & $\bullet$ \\
\cite{LenoD3,LenoD4,LenoD2,LenoD1,Leno2019} & $\bullet$ & $\bullet$ & $ \bullet$ & $\bullet$ & $\bullet$ & $\bullet$ \\
\cite{Linares2015} & $\bullet$ & $\bullet$ & $\bullet$ & single & & $\bullet$ \\
\cite{Linn2018} & $\bullet$ & $\bullet$ & & $\bullet$ & & $\bullet$ \\
\cite{Urabe2021} & & $\bullet$ & $\bullet$ & $\bullet$ & $\bullet$ & $\bullet$ \\
\bottomrule
\end{tabularx}
\label{tab:litreview}
\end{table}

\begin{enumerate}
    \item An \emph{action type}, which describes the action a user takes. Actions are most often divided into mouse and keyboard inputs, but some logs further distinguish between different mouse buttons, string inputs, and hotkeys. Only two logs do not record action types: the BPIC 2016 clickstream log \cite{BPIC2016} and Urabe et al. \cite{Urabe2021}, who only record that an interaction has taken place.
    \item The atomic \emph{target UI element}, on which the user action is executed.
    This attribute is recorded in most UI logs, except for two: Dev et al. \cite{Dev2017} only record the usage of specific functions, such as crop in a graphics editor, and Jimenez-Ramirez et al. \cite{Jimenez2019,JimenezD1} record click coordinates and screenshots, but only use them to match similar user actions and do not map them to target elements.
    \item The \emph{software application} that the user interacts with. This could be a web browser, an ERP system, or an office application. This attribute is always recorded when researchers track user actions across multiple applications, but is not captured when the tracking is limited to a single application.
    \item One or multiple attributes that specify the location of the target element in the application's \emph{UI hierarchy}. For example, an Excel cell is located in a worksheet (hierarchy level 1), which belongs to a workbook (hierarchy level 2) \cite{Leno2019}. UI hierarchy attributes are included in about half the reviewed logs.
    \item The \emph{input value} that the user writes into a text field. Input values are included in about half of the reviewed logs.
    \item A \emph{timestamp}, which records the exact date and time at which the action occurred. This is recorded in all logs. 
\end{enumerate}

\noindent
\textbf{Differences.}
Most authors characterize user interactions through an action type, i.e., \emph{what} the user does, and a target element, i.e., \emph{where} they do it. 
However, the set of possible values for the action type, and hence the level of detail at which actions are recorded, differs considerably. For example, Agostinelli et al. \cite{Agostinelli2020} record aggregated action types abstracted from raw hardware input (e.g., clickButton and clickTextField), whereas Jimenez-Ramirez et al. \cite{Jimenez2019} make the low-level differentiation between left, right, and middle mouse clicks. 
Which other attributes are included in a UI log differs between approaches: whereas timestamps and the application in focus (where applicable) are recorded in all logs, input values and information on the location of a target element within the application's UI hierarchy are included in about half of them. Examples for other, less common attributes that are only recorded in few approaches include the current value of a text field \cite{Leno2019,Choi2021,AgostinelliD1}, user IDs \cite{Urabe2021,AgostinelliD1,Leno2019,Damevski2017}, other resources involved \cite{Ardimento2019,BPIC2016}, and associations to higher-level process steps \cite{Linn2018, Hofmann2021}.

Another interesting finding was that most of the reviewed UI logs are initially unlabeled, i.e., they do not have a concrete case notion \cite{Ferreira2009}. In some publications, events in unlabelled logs are later grouped into cases based on different attributes. These attributes include external session IDs created automatically by a system \cite{BPIC2016} or manually by users \cite{Leno2019,Agostinelli2020,Linares2015}, user IDs \cite{Damevski2017}, or case IDs from associated higher-level event logs \cite{Hofmann2021}.

\section{Review of Industry Solutions} 
\label{sec:industryreview}

To ensure broad applicability of our reference data model, we also review industry approaches for conceptualizing and capturing UI logs. 
Because those approaches are core to the industry solutions' functionality and business secrets, the available material for this review may be less specific than scientific papers. 
Therefore, we conduct the industry review in this section as an addition to the literature review, meant to confirm and complement the established findings. 


\subsection{Research Method}

\noindent
\textbf{Selection Strategy.}
An initial analysis indicated that RPA tools are presently the only industry solutions that collect UI logs on a large scale.
Some vendors also advertise task mining capabilities, but their primary focus is on recording UI logs for the automation of routines. 
Because the RPA market is highly fractured and fast-moving, we could not conduct a complete review. 
Instead, we opted to analyze a sample of companies that can be seen as representative for the market.
Therefore, we selected the companies that the 2021 Gartner Magic Quadrant RPA report\footnote{\url{https://www.gartner.com/en/documents/3988021}} attributes with a ``high ability to execute'' and/or a ``high completeness of vision'':
UiPath, Automation Anywhere, Microsoft Power Automate, Blue Prism, NICE, WorkFusion, Pegasystems, Appian, EdgeVerve Systems, and Servicetrace. 
We also included Celonis Task Mining, which is the only major product that uses UI logs primarily for low-level process mining.

\noindent
\textbf{Review Approach.}
In analyzing those eleven tools, we focused on finding the commonalities and differences between the industry logs and the scientific logs. Specifically, we wanted to know whether the industry logs capture the same set of six core attributes found in the scientific logs (commonalities) and whether the industry logs capture any other attributes that could be relevant for a widely applicable reference data model (differences). 
To answer those questions, we collected freely available material about the tools.\footnote{A full list of the material that we analyzed can be found at \url{https://gitlab.uni-mannheim.de/jpmac/ui-log-data-model/-/blob/46b363dc75b992a43398f501e3a4cb0e755107d0/industry_review_sources.pdf}} 
This included trial or demo versions, documentations, and promotional material, such as videos showcasing the recording process.
After collecting the material, we had to exclude two companies from our list, Pegasystems and EdgeVerve Systems, because we could not obtain sufficient information on the functionalities of their recording software. 

\subsection{Results}

\noindent
\textbf{Commonalities.}
For each industry solution, we analyzed whether it also records the six core attributes found in the literature review. The results are summarized in \autoref{tab:industryreview}. All reviewed tools record action types, target elements, input values, applications, and timestamps. Similar to what we found in the literature review, the recordable action types differ considerably between tools.

\begin{table}[hbt]
\vspace{-2em}
\caption{UI log attributes as found in the industry review}
\begin{tabularx}{\textwidth}{lYYYYYY}
\toprule
\textbf{Company} & \textbf{Action type} & \textbf{Target element} & \textbf{UI Hierarchy} & \textbf{Appli-cation} & \textbf{Input value} & \textbf{Time-stamp} \\
\midrule
UiPath & $\bullet$ & $\bullet$ & screenshots & $\bullet$ & $\bullet$ & $\bullet$ \\
MS Power Automate & $\bullet$ & $\bullet$ & $\bullet$ & $\bullet$ & $\bullet$ & $\bullet$ \\
Automation Anywhere & $\bullet$ & $\bullet$ & $\bullet$ & $\bullet$ & $\bullet$ & $\bullet$ \\
Celonis & $\bullet$ & $\bullet$ & screenshots & $\bullet$ & $\bullet$ & $\bullet$ \\
Blue Prism & $\bullet$ & $\bullet$ & screenshots & $\bullet$ & $\bullet$ & $\bullet$ \\
Workfusion & $\bullet$ & $\bullet$ & screenshots & $\bullet$ & $\bullet$ & $\bullet$ \\
NICE & $\bullet$ & $\bullet$ & screenshots & $\bullet$ & $\bullet$ & $\bullet$ \\
Appian & $\bullet$ & $\bullet$ & screenshots & $\bullet$ & $\bullet$ & $\bullet$ \\
ServiceTrace & $\bullet$ & $\bullet$ & screenshots & $\bullet$ & $\bullet$ & $\bullet$ \\
\bottomrule
\end{tabularx}
\label{tab:industryreview}
\end{table}

\noindent
\textbf{Differences.}
We also examined whether the industry solutions systematically record any other attributes, but we did not find any. 
However, we did find a significant difference between industry logs and scientific logs in how they capture information on the location of elements within the UI hierarchy. In research, this information is explicitly recorded in UI log attributes, but most industry tools instead store it as screenshots outside of the log. Some tools also use the UI hierarchy to construct selectors that uniquely identify an element within an application's GUI, similar to file paths.
Another difference between industry logs and scientific logs concerns the case notion. In the industry solutions, the case ID is always a task or process label that is manually added to the log. Additional business context attributes also need to be added by users and are not recorded by the tool.

\section{Reference Data Model} 
\label{sec:data_model}

In this section, we introduce our reference data model for user interactions. 
We consider a reference model to be a conceptual model that serves to be reused for the design of other conceptual models \cite{rehse2019procedure}. 
Under such a reuse-oriented conceptualization, (universal) applicability of the model is not a defining property.
However, maximizing the model's application scope increases its reuse potential and therefore its value to the community. 
Therefore, we designed the model in an inductive or bottom-up fashion \cite{rehse2019procedure}: based on the commonalities and differences between existing UI logs that we found in the literature and industry reviews, we constructed a model that subsumes those commonalities, but remains flexible with regard to their differences. 
In the following, we first elaborate on the principles that guided our design process in \autoref{subsec:principles}. The reference model is presented in detail in \autoref{subsec:model}. In \autoref{subsec:exchange}, we provide a data interchange format for UI log data as a supplement to the reference model.

\subsection{Design Principles}
\label{subsec:principles}

In the literature and industry reviews, we found that the main commonality between existing UI logs are the six core attributes. The main differences between them concerned the scope, the level of abstraction, and the case notion.
Based on these findings, our data model follows four fundamental design principles:

\begin{enumerate}
    \item \textbf{Minimal set of core components}: The essential characteristics of user interactions, as found in the reviews, are modeled as the components and standard attributes of the data model. Because the model is intended to be non-specific and universally applicable, we include no other elements, thus keeping the number of components and standard attributes to a minimum.
    \item \textbf{Flexible scope}: To ensure flexibility in scope, the data model can be extended with any number of additional components and all components can have an arbitrary number of attributes. Also, nearly all components and standard attributes are optional. The only non-optional component and attribute that ensure the existence of a UI log are the activity and its name.
    \item \textbf{Flexible level of abstraction}: To enable user interactions to be modeled in various application contexts and at various levels of abstraction, the domain of the standard attributes in the data model, such as the action type, is left unspecified and can be determined at the point of instantiation.
    Furthermore, all components are modeled as classes and can be subclassed. Explicit subclasses are only defined for the target object, because they are inherent to the structure of user interfaces and the way they are embedded in ISs.
    \item \textbf{No explicit case notion}: Whereas the case notion of a business processes is tied to its instances, UI logs are not inherently structured along any data dimension. The reviews have shown that they can have many possible case identifiers. The data model therefore does not include an explicit case notion. Instead, the case notion needs to be defined at the point of instantiation. 
\end{enumerate}

\subsection{Reference Model Components}
\label{subsec:model}

\begin{wrapfigure}{L}{0.65\textwidth} 
\centering
\vspace{-5mm}
\includegraphics[width=0.6\textwidth] 
{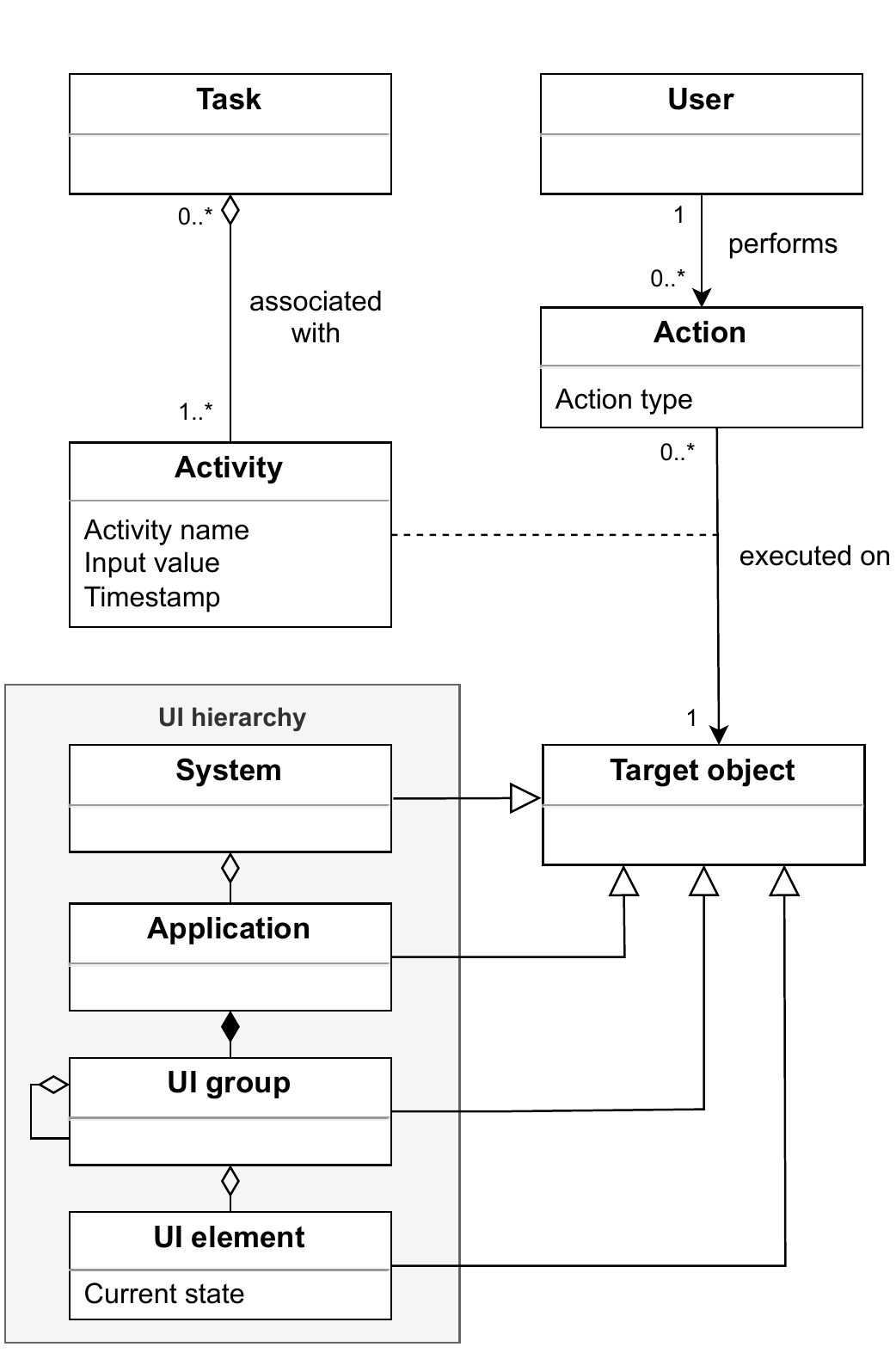}
\caption{User interaction data model}
\label{fig:datamodel}
\vspace{-5mm}
\end{wrapfigure}

The reference data model is depicted as a UML diagram in \autoref{fig:datamodel}.  
It consists of nine components, modeled as classes, and their interrelations, modeled as associations.
Each class has an ID and can have any number of attributes.
Some components have standard attributes that have a particular significance for user interactions.
In the following, we define and explain the individual components. 

\noindent
\textbf{Components that define the user interaction.}
In our model, user interactions have two parts. First, the \emph{action} component with its \emph{action type} standard attribute that describes what the user does. Common action types, as observed in the reviews, correspond to the functionalities of standard peripheral input devices, such as left or right mouse clicks, single keystrokes, or keystroke combinations for shortcuts. Higher-level distinctions are also possible. For example, when collecting data in an ERP system, actions can be divided into input actions, which make changes to a business object, and navigation actions, which only serve to navigate the GUI. 

The second part of an interaction is the \emph{target object} that the action is executed on. It is instantiated as one of four object types in the UI hierarchy, as explained below.
The action type and target object together determine the central model component: the \emph{activity}. It is uniquely defined as a combination of an action and a target object and acts as the event label, like in a traditional event log. An activity has three standard attributes: the \emph{activity name}, an optional \emph{input value} that denotes, e.g., the string that is entered into a text field, and a \emph{timestamp} to indicate its execution time. The activity name is determined as a function of the action type of the corresponding action and the identifier of the corresponding target object, for example, a concatenation. The timestamp is a very common attribute in traditional event logs as well as UI logs. It is, however, not a strictly required attribute in the data model, since there are alternative ways to introduce a notion of order into an event log \cite{order}.

\raggedbottom

\noindent
\textbf{Components that define the UI hierarchy.}
The UI hierarchy integrates the various types of UI element context data into a general structure. It consists of four components, which form a tree-shaped composition hierarchy: UI element, UI group, application, and system. The UI element and UI group levels mirror the hierarchical structure of virtually all GUIs (e.g., the document object model of a website). The application and system levels go beyond the actual GUI and position it within an IS, which makes it possible to record application- and system-level user interactions and allows the UI log to be compatible with cross-application and even cross-system UI tracking.

Most actions are executed on the atomic \emph{UI elements}, which form the lowest level. Examples include buttons, text boxes, dropdowns, checkboxes, or sliders. Elements can be stateful, such as a non-empty text box or a greyed-out button. Capturing this state is necessary, for example, to track the effects of copy/paste actions or to differentiate between activity outcomes. The state of a UI element is therefore recorded in its \emph{current state} standard attribute.

UI elements are combined into \emph{UI groups}, which can be nested within other UI groups. In many cases, these UI groups are explicit design elements of the user interface, but our model does not impose grouping criteria and allows UI groups to be formed from arbitrary sets of UI elements. A simple example that we saw in the literature review is an Excel cell (UI element), which is part of a worksheet (UI group), which is again part of a workbook (UI group).
Modeling UI groups has two main advantages. First, it allows to uniquely identify functionally identical UI elements. For the example above, recording information about UI groups allows us to distinguish between the cell A1 in separate Excel worksheets. This idea is used in many industry solutions to generate element selectors from screen captures. Second, UI groups can be useful for event abstraction, i.e., mapping user interactions to higher-level conceptual tasks, if these tasks are closely tied to particular UI groups. For example, all interactions with elements in a login mask (enter username, enter password, click login) can directly be abstracted to the ``login'' task. 

UI elements and UI groups belong to an \emph{application}, i.e., a single program instance. Some actions are directly executed on the application and are not tied to lower-level elements, such as ``undo'' or application-specific hotkeys.

The root node of the UI hierarchy is the \emph{system}, on which the applications run and actions are recorded. Similar to application-level actions, it is also possible to capture system-level actions, such as
the Ctrl-Alt-Del key combination to open the Task Manager on a Windows system. 

\noindent
\textbf{Components that define the context.}
Finally, the data model includes two components that put UIs in a conceptual context: \emph{user} and \emph{task}. These exist in some form for all UI logs, which is why they are included in the model. In contrast, other potential context components, such as organizational or resource attributes, are use-case-specific and can be considered by extending the model. 

The user is the entity that initiates any interaction. Each action is associated with a single user. Because user IDs and attributes depend on the data collection environment (e.g., device IDs in mobile applications or IP addresses on websites), the model does not specify any attributes for users. This also means that the user component is not necessarily restricted to humans and can model computer-initiated interactions, for example when recording partially automated processes.

The task component associates the recorded user interactions with conceptual tasks or routines, which makes it possible to map low-level GUI interactions to higher-level user activities. This abstraction is an essential prerequisite for being able to perform meaningful analysis on UI logs or to use them for automation.

\subsection{Exchange Format} 
\label{subsec:exchange}

To further increase the applicability and reuse potential of our data model, we implemented it as an extension to the XES standard for event logs.\footnote{The XML specification for the \emph{UILog} extension is available at \url{https://gitlab.uni-mannheim.de/jpmac/ui-log-data-model/-/raw/main/UILog_extension}.} 
This \emph{UIlog} extension provides a standardized exchange format for UI logs as a supplement to the data model. 
The implementation considers the activity equivalent to the event label and does not include the activity name or timestamp standard attributes because those are already provided by the concept and time extensions. The other components and standard attributes are defined at event level, i.e., as attributes of an activity instance. The generic target object is not directly implemented, but can instead be specified through attributes that correspond to its four UI hierarchy subclasses: the target object is the lowest-level UI hierarchy component that exists for this event. For example, for an event with a UI element attribute, the target object is always this UI element, whereas for an event with no UI element or UI group attributes, the target object is the application.

\section{Working Example} 
\label{sec:evaluation}

To demonstrate the practical utility of the reference data model, we describe in this section how it can be instantiated in a real-life scenario.
The scenario is based on an RPA project, which we are currently conducting in cooperation with an ERP system vendor.
The project is set in the medical technology industry, where companies are required to regularly validate their ISs to ensure that they are in compliance with external quality regulations. The validation of an IS involves manually executing a number of predefined workflows step-by-step according to a rigid execution plan, checking the result of each step against a set of acceptance criteria, and documenting the result.
Manually executing a well-defined validation workflow is a repetitive and time-consuming task. 
The goal of our project is to automate this task using RPA. 
We want to record how process experts interact with the UI of the ERP system during validation and then train bots to emulate their actions. 

\begin{figure}[t!]
\centering
    \begin{minipage}{0.3\textwidth}
        \centering
        \includegraphics[width=\textwidth]{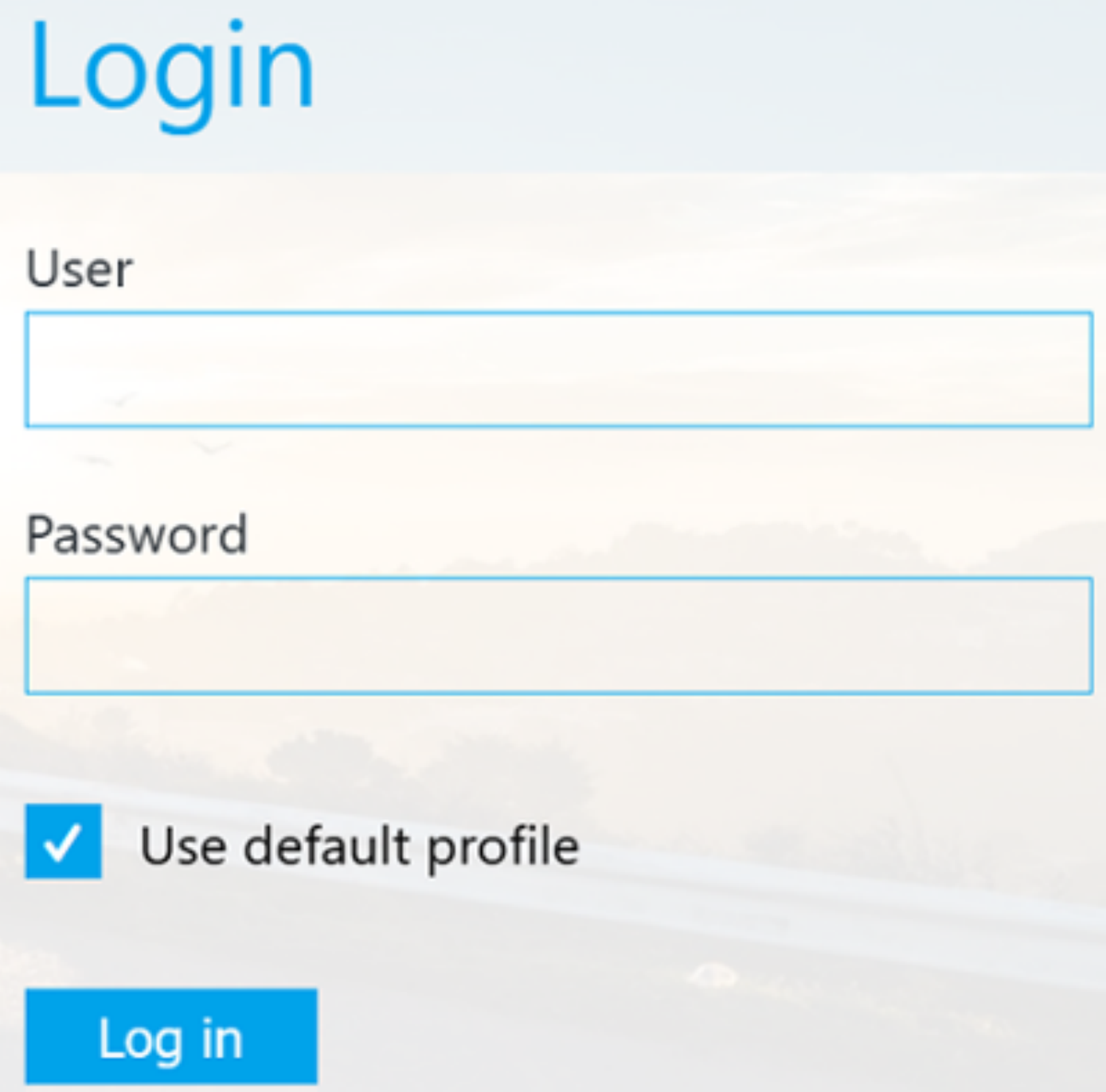}
        a) Login
    \end{minipage} \hspace{1cm}
    \begin{minipage}{0.3\textwidth}
        \centering
        \includegraphics[width=\textwidth]{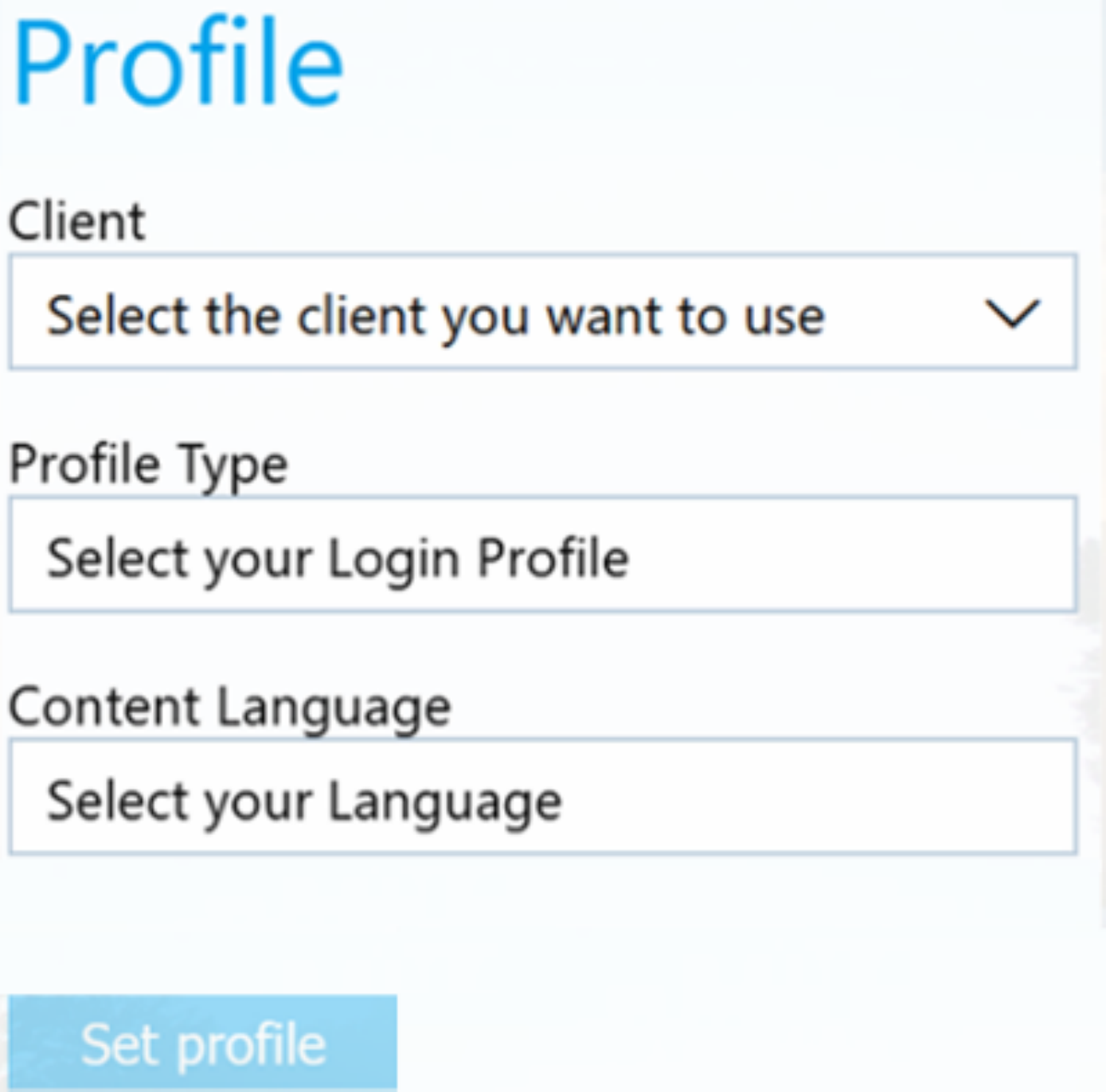}
        b) Profile selection
    \end{minipage}
    
    \begin{minipage}{0.7\textwidth}
        \centering
        \includegraphics[width=\textwidth]{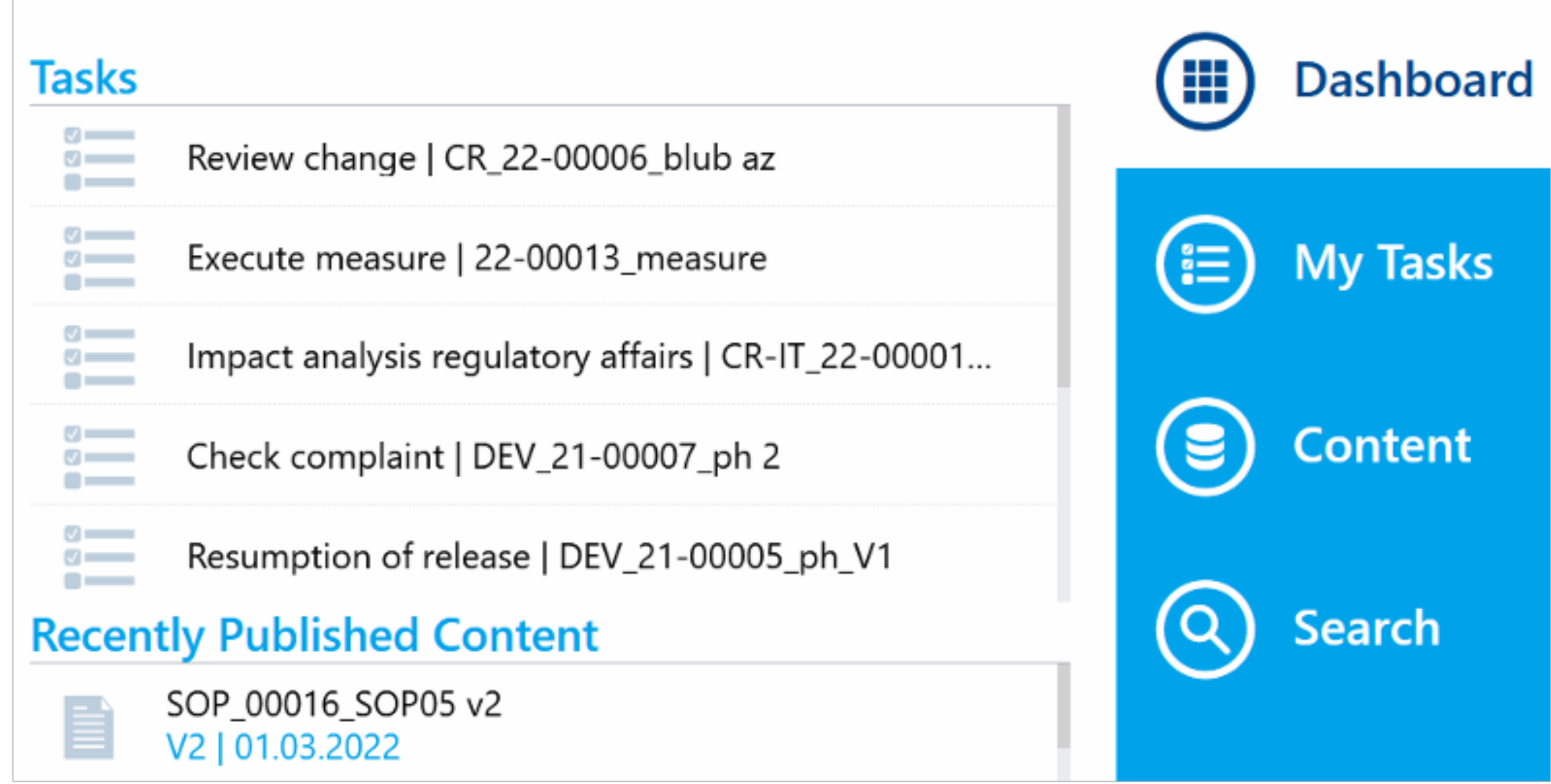}
        c) Dashboard
    \end{minipage}
    \vspace{\abovedisplayskip}
    
    \begin{minipage}{0.8\textwidth}
        \centering
        \includegraphics[width=\textwidth]{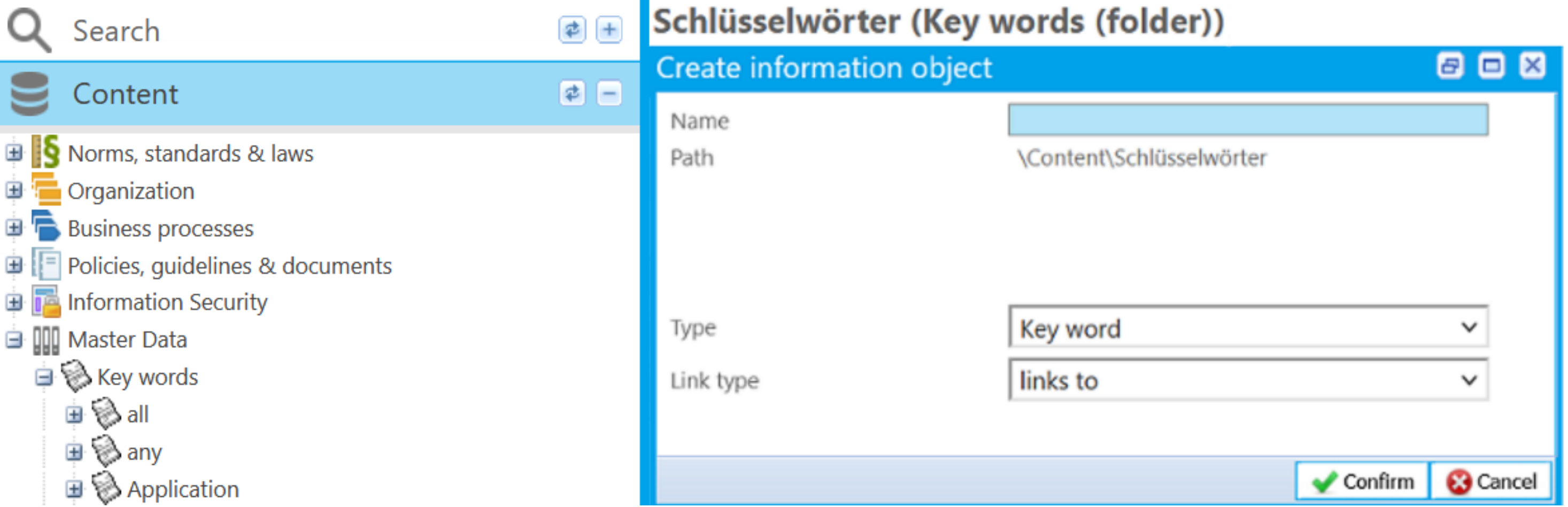}
        d) Explorer tree with "Create information object" form panel
    \end{minipage}
\caption{The user interface of the application described in the example}
\label{fig:dvision_ui}
\vspace{-1.5em}
\end{figure}

In the following, we use the example of a keyword creation workflow to
show how an artificial UI log that captures one execution of this workflow may instantiate the data model.
The keyword creation workflow consists of five consecutive steps, which are executed on the GUI parts shown in \autoref{fig:dvision_ui}.
The user 
\begin{inparaenum}[(1)]
\item logs in (a), 
\item selects the right client and profile (b), 
\item navigates through the dashboard (c) to reach the explorer tree (d, left), \item creates a new keyword (d, right), and 
\item logs out. 
\end{inparaenum}
The main acceptance criterion is that the newly created keyword shows up in the explorer tree after refreshing. 

\autoref{tab:logexample} shows a UI log for one case, i.e., one execution of the keyword creation workflow. 
It includes the action type, target UI element, and one level of UI groups, plus input value and current state where applicable. 
The captured action types are left and right clicks, text input, selected keyboard shortcuts, and none. The activity label of an event (in most cases) consists of the concatenated action type and target object identifier. 

\begin{table}[htb]
\scriptsize
\centering
\RaggedRight
\caption{UI log for one execution of the keyword creation workflow}
\begin{tabularx}{\textwidth}{p{2cm}lp{2cm}lp{2cm}p{1.9cm}}
\hline
\textbf{Activity} & \textbf{Action type} & \textbf{UI element} & \textbf{UI group} & \textbf{Input value} & \textbf{Current state} \\ 
\hline
A\_Login & none &  & login mask & \{username: pren, password: dts123\} &  \\ \hline
A\_Profile Selection & none &  & user select client & \{client: base, profile: author\} & \\ \hline
click content & left click & content & dashboard ov &  & \\ \hline
click masterdata & left click & masterdata & explorer tree &  & \\ \hline
click masterdata node expand & left click & masterdata node expand & explorer tree &  & \\ \hline
click keywords node expand & left click & keywords node expand & explorer tree &  & \\ \hline
rclick keywords & right click & keywords & explorer tree &  & \\ \hline
click ppanel new & left click & ppanel new & explorer tree &  & \\ \hline
click new information object & left click & new information object & explorer tree &  & \\ \hline
click name & left click & name & fpanel keyword &  & \\ \hline
input name & input & name & fpanel keyword & MyKeyword & \\ \hline
click dd type & left click & dd type & fpanel keyword &  & [keyword, keywords folder] \\ \hline
click dd type & left click & dd type & fpanel keyword & keyword & [keyword, keywords folder] \\ \hline
click dd linksto & left click & dd linksto & fpanel keyword & & [linksto] \\ \hline
click dd linksto & left click & dd linksto & fpanel keyword & linksto & [linksto] \\ \hline
click confirm & left click & confirm & fpanel keyword & & \\ \hline
click keywords node expand & left click & keywords node expand & explorer tree &  & \\ \hline
KEY\_F5 explorer tree & KEY\_F5 &  & explorer tree &  & \\ \hline
click logout & left click & logout & explorer tree &  & \\ \hline
click confirm & left click & confirm & dialog logout &  & \\
\end{tabularx}
\label{tab:logexample}
\vspace{-2em}
\end{table}

The first two events in the log do not correspond to single user interactions, but instead take advantage of the UI group concept to directly abstract to higher-level tasks. Instead of recording each event in the login and client selection masks separately, the task is tracked only at completion and the content of the text fields is read out when the user presses the ``Login'' and ``Set Profile'' buttons. For these abstracted activities (marked with an ``A\_'' prefix), the action type is ``none''; they are defined only through the target UI group, independent of the performed actions. This approach can be used for simple tasks with the same execution pattern in all workflows.
Its main upside is that it reduces noise, which is a common problem in UI logs \cite{LenoD1}. In our scenario, activities like initially entering a wrong password do not affect the outcome of the workflow and are therefore not relevant for automating it. 
By abstracting during data collection, those activities are automatically disregarded. Other advantages of abstraction are reduced implementation effort and smaller UI logs.

For effective automation, various user inputs need to be tracked. Therefore, the instantiation of the input value attribute in the log is flexible and depends on the action type and target object: When a user writes into a textbox, the input value is the entered string. When an item is selected from a dropdown, the input value records the label of that item. For abstracted activities, the input value captures the string values of all relevant UI group elements as a map.
Most state information, however, is not required for automation. Therefore, the current state attribute only records the values that can be selected from list and dropdown elements, which is needed for some more complex workflows in the validation process. For example, if a documents needs to be approved, the validation must verify that a document's author cannot be selected as approver.

This simple example demonstrates how some of the core components of the reference model can be instantiated in a real-life scenario, and how the flexibility in abstraction level can be leveraged to record attributes in a way that matches the requirements of a particular use case. It also shows that, in practice, components that are not relevant for a use case can simply be left out. The main advantage of using the reference model here is that, unlike with an ad-hoc model tailored to the use case, the attributes captured in the UI log follow a general convention that also applies to other user interfaces. This makes recording UI logs in the same format straightforward even in other applications, and makes it possible to develop automation or task mining solutions that are independent of the recording approach used.


\section{Discussion and Conclusion}
\label{sec:discussion}

In this paper, we propose a reference data model for UI logs. Based on reviews of scientific literature and industry solutions, it has a set of core components to capture essential characteristics of user interactions and is flexible with regard to scope, abstraction level, and case notion. We implement the model as an XES extension and exemplarily show how it can be instantiated in a real-life RPA scenario.


\noindent
\textbf{Contribution.} 
Our main objective is to address the issues that arise from the lack of standardization of UI logs. Therefore, we derive the reference from existing UI logs. Most of the components 
are directly adopted from the core attributes identified in our reviews. Our contribution is their integration into a unified framework with well-defined relations. For example, we propose a rigid interpretation of an activity by defining it as a combination of an action and a target element. We also expand on the location context of UI elements 
and explicitly define four distinct types of target objects in an unambiguous hierarchy.

To this unified framework, we add additional, less frequently collected components and standard attributes that are particularly relevant for a complete model of user interactions. For instance, the current state is an important property of stateful UI elements when the log is intended to be used for automation. In the UI hierarchy, we add the system on top of the commonly recorded application to model system-level user interactions. We also introduce the user and task context components to add (optional) generic business context to UI logs.

\noindent
\textbf{Limitations.} 
One limitation of our work concerns its grounding in existing UI logs. Despite following a methodical approach, we do not claim that our reviews or the model are complete or exhaustive. There could be unidentified UI logs or future UI logs in different use cases, which are not well represented by the model. For instance, our data model is only intended to model user interactions with graphical user interfaces, and we did not consider alternative input types, for example from voice commands or eye-tracking devices. The model may also be somewhat biased towards automation use cases because RPA solutions are overrepresented in the two reviews that it is based on.

Another limitation is that the XES standard is not particularly well-suited for UI logs. It does not support explicitly defining the relations between attributes, so all components of the UI hierarchy have to be implemented at event level. Therefore, even if many events involve the same target object, UI group, application, system and their attributes need to be included each time, leading to considerable redundancy. XES also assumes a single case notion, contrary to the flexible case notion that we intend for the data model.

Conceptually, implementing the model as an extension to the Object-Centric Event Log format OCEL \cite{OCEL} would be more appealing, because users, tasks, and UI hierarchy elements could be modeled as objects, reducing the redundancy. However, OCEL does not support extensions and has two main limitations with regard to UI logs. First, it does not support dynamic object attributes that can differ between events, such as the current value of a textbox that may change between interactions. Second, object attributes cannot be tied to certain object types, so for example the current state attribute cannot be limited to UI elements only.
Therefore, we decided to implement the model as an XES extension.

\noindent
\textbf{Future Work.} 
Our reference model can contribute to the field by providing a common, application-independent conceptual framework for user interactions. However, like any reference model, it needs to prove its utility in practice. We therefore want to encourage researchers and practitioners to adopt the model for capturing UI logs in their projects, and to extend it both with regard to new use cases and with regard to conceptual aspects, such as user privacy.

\bibliographystyle{splncs04}
\bibliography{bibliography}

\end{document}